\pdfoutput=1
\newcommand{\CLASSINPUTtoptextmargin}{54pt}      % top text margin
\newcommand{\CLASSINPUTbottomtextmargin}{54pt}   % bottom text margin
\newcommand{\CLASSINPUTinnersidemargin}{54pt}    % inner side margin
\newcommand{\CLASSINPUToutersidemargin}{54pt}    % outer side margin

\documentclass[letterpaper, conference]{IEEEtran}
\IEEEoverridecommandlockouts
\usepackage[caption=false,font=footnotesize]{subfig}
\usepackage[pdfauthor={Sachithra Hemachandra, Felix Duvallet, Thomas
  M. Howard, Nicholas Roy, Anthony Stentz, Matthew R.Walter},%
pdftitle={Learning Models for Following Natural Language Directions in Unknown Environments}]{hyperref}
\usepackage{amsmath}
\usepackage{amsfonts}
\usepackage{amssymb}
\usepackage{graphicx}
\usepackage{tikz}
\usepackage{pgfplots}
\pgfplotsset{compat=1.9}
\usepgfplotslibrary{statistics}
\usepackage{tabularx}
\usepackage{etex}
\usepackage{booktabs}
\usepackage[ruled]{algorithm2e}

\usepackage[nameinlink, capitalise]{cleveref}

\newcommand{\argmin}[1]{\ensuremath{\underset{#1}{\arg \! \min}\;}}
\newcommand{\argmax}[1]{\ensuremath{\underset{#1}{\arg \! \max}\;}}

\newcommand*{\parens}[1]{\left( #1 \right)}

\newcommand*{\brackets}[1]{ \left[ #1 \right] }

\newcommand*{\Wtranspose}{W^T}
\newcommand*{\witranspose}{w_i^T}
\newcommand*{\astar}{{a^{*}}}
\newcommand*{\pistar}{\pi^*}

\newcommand*{\dldwfrac}{\frac{\partial \ell}{\partial W}}

\newcommand*{\Mapt}{S_t}
\newcommand*{\Mapti}{S_t^{\parens{i}}}
\newcommand*{\Momenta}{\hat{\Phi}_1}  % Cannot use numbers.
\newcommand*{\Momentb}{\hat{\Phi}_2}
\newcommand*{\Momenti}{\hat{\Phi}_i}
\newcommand*{\Momentk}{\hat{\Phi}_k}

\newcommand*{\XAMapt}{\parens{x, a, \Mapt}}

\newcommand*{\phiXAMapi}{\phi \parens{x, a, \Mapti}}

\newcommand*{\costMapt}{c\parens{x, a, \Mapt}}
\newcommand*{\costMaptAstar}{c\parens{x, \astar \negmedspace, \Mapt}}

\newcommand*{\astop}{a_\textrm{stop}}
\newcommand*{\visited}{\mathcal{V}}
\newcommand*{\frontiers}{\mathcal{F}}
\newcommand{\DAgger}{\textsc{DAgger}}

\definecolor{blue}{rgb}{0,.0,1} %\newcommand{\fr}{\color{blue}}
\definecolor{orange}{rgb}{1,0.5,0}
\definecolor{fullred}{rgb}{0.85,.0,.1}

\newcommand{\postTableSpace}{\vspace{-0pt}}

\definecolor{colorNLMMP}{rgb}{0.5, 0.5, 0.5}
\definecolor{colorBelief}{rgb}{0.7,0.7,0.7}

\title{\vspace{6mm} \LARGE \bf
Learning Models for Following Natural Language Directions \\in Unknown Environments
}

\author{Sachithra Hemachandra$^*$ \quad Felix Duvallet$^*$ \quad%
  Thomas M.~Howard\\ Nicholas Roy \quad Anthony Stentz \quad Matthew R.~Walter%
\thanks{$^*$The first two authors contributed equally to this
  paper.}%
\thanks{S.~Hemachandra and N.~Roy are with the
  Computer Science and Artificial Intelligence Laboratory,
  Massachusetts Institute of Technology, Cambridge, MA USA {\tt\small
    \{sachih,tmhoward,nickroy\}@csail.mit.edu}}%
\thanks{F.~Duvallet and A.~Stentz are with the Robotics
  Institute, Carnegie Mellon University, Pittsburgh, PA USA {\tt\small
  \{felixd,tony\}@cmu.edu}}%
\thanks{T.M.~Howard is with the University of Rochester, Rochester, NY USA {\tt\small thomas.howard@rochester.edu}}
\thanks{M.R.~Walter is with the Toyota Technological Institute at
  Chicago, Chicago, IL USA {\tt\small mwalter@ttic.edu}}
}

\begin{document}

\maketitle

% -*- mode:LaTeX; mode:visual-line; mode:flyspell; fill-column:75 -*-

\begin{abstract}
    Natural language offers an intuitive and flexible means for humans to
    communicate with the robots that we will increasingly work alongside in our homes and
    workplaces. Recent advancements have given rise to robots that are
    able to interpret natural language manipulation and navigation
    commands, but these methods require a prior map of the robot's environment. In
    this paper, we propose a novel learning framework that enables robots to
    successfully follow natural language route directions without any
    previous knowledge of the environment. The algorithm utilizes spatial
    and semantic information that the human conveys through the command to
    learn a distribution over the metric and semantic
    properties of spatially extended environments. Our method uses this
    distribution in place of the latent world model and interprets the
    natural language instruction as a distribution over the intended
    behavior. A novel belief space planner reasons directly over the map
    and behavior distributions to solve for a policy using imitation
    learning. We evaluate our framework on a
    voice-commandable wheelchair. The results demonstrate that by
    learning and performing inference over a latent environment model, the
    algorithm is able to successfully follow natural language route
    directions within novel, extended environments.
\end{abstract}

% -*- mode:LaTeX; mode:visual-line; mode:flyspell; fill-column:75 -*-

\section{Introduction} \label{sec:introduction}

Over the past decade, robots have moved out of controlled isolation
and into our homes and workplaces, where they coexist with people in
domains that include healthcare and manufacturing. One long-standing
challenge to realizing robots that behave effectively as our partners
is to develop command and control mechanisms that are both intuitive
and efficient. Natural language offers a flexible medium
through which people can communicate with robots, without requiring
specialized interfaces or significant prior
training. For example, a voice-commandable
wheelchair~\cite{hemachandra11} allows the mobility-impaired to
independently and safely navigate their surroundings simply by speaking to
the chair, without the need for traditional head-actuated switches or
sip-and-puff arrays. Recognizing these advantages, much attention has been paid
of late to developing algorithms that enable robots to interpret natural
language expressions that provide route directions~\cite{macmahon06,
  kollar10, chen11, matuszek12}, that command manipulation~\cite{tellex11,
  howard14a}, and that convey environment knowledge~\cite{walter13,
  hemachandra14}.
\begin{figure}[!t]
    \centering
    \includegraphics[width=1.0\linewidth]{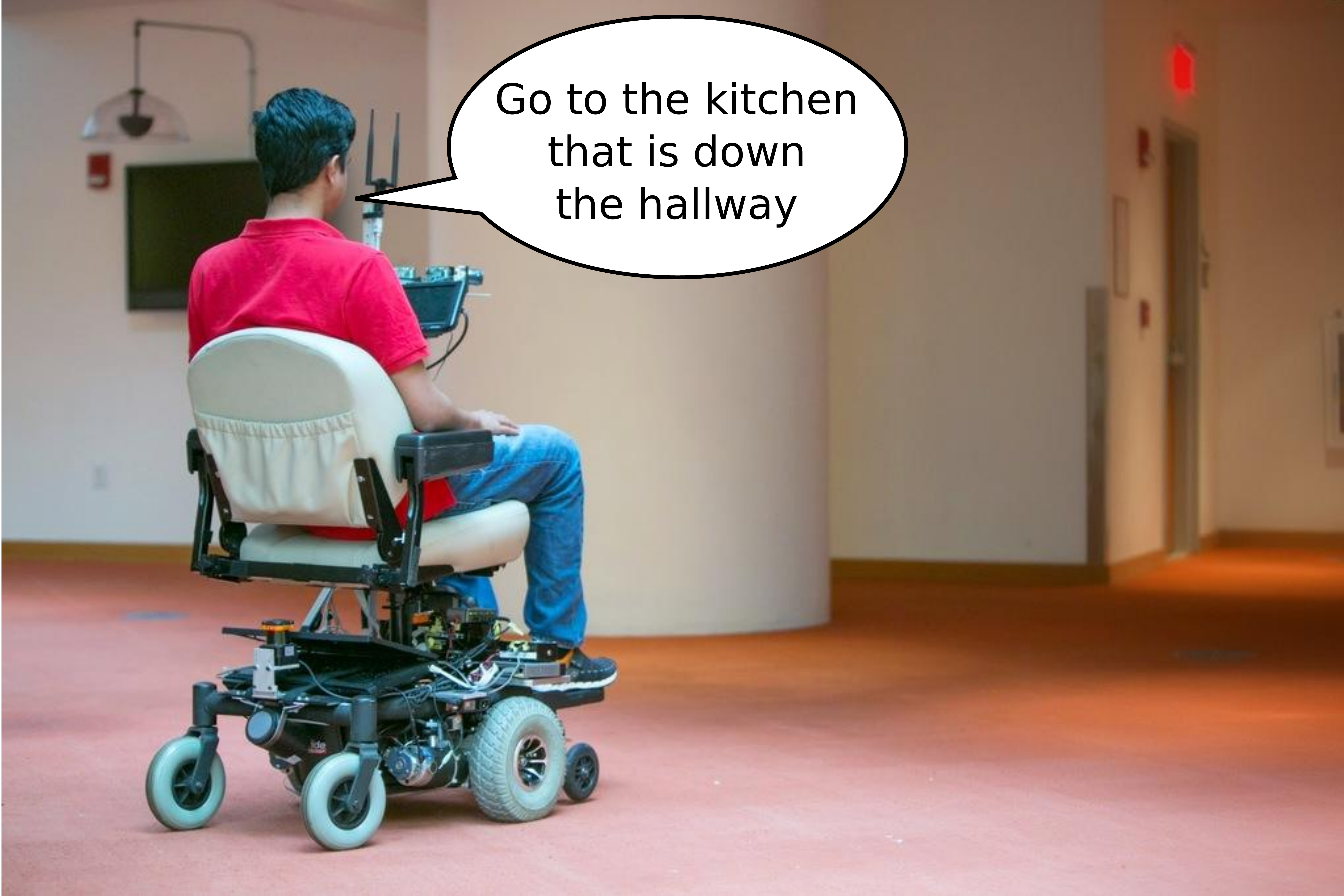}
    \caption{Our goal is to enable robots to autonomously follow natural
      language commands without any prior knowledge of their environment.}
    \label{fig:go_to_kitchen_down_hallway}
\end{figure}
%
%

% Put this here for pagination.
% -*- mode:LaTeX; mode:visual-line; mode:flyspell; fill-column:75 -*-
\newcommand{\includeValueGraphic}[2][]{
  \includegraphics[
    angle=0,
    % left bottom right top
    trim = 25mm 10mm 10mm 20mm, clip,
    width = 0.315\textwidth,
    #1
  ]{#2}
}
\begin{figure*}[!tb]
  \centering
  \subfloat[][\label{fig:value1}$t=3$]{%
    \includeValueGraphic{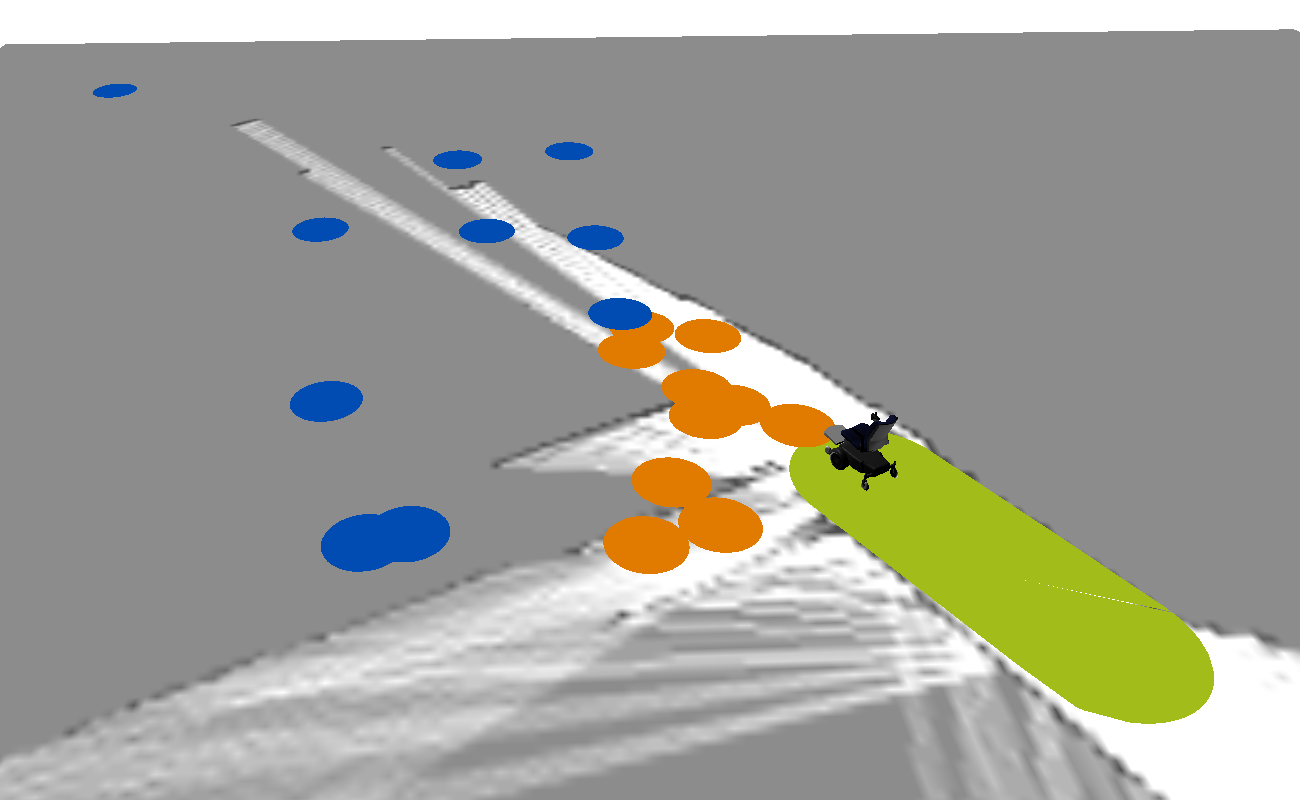}%
  }\hfill%
  \subfloat[][\label{fig:value2}$t=4$]{%
    \includeValueGraphic{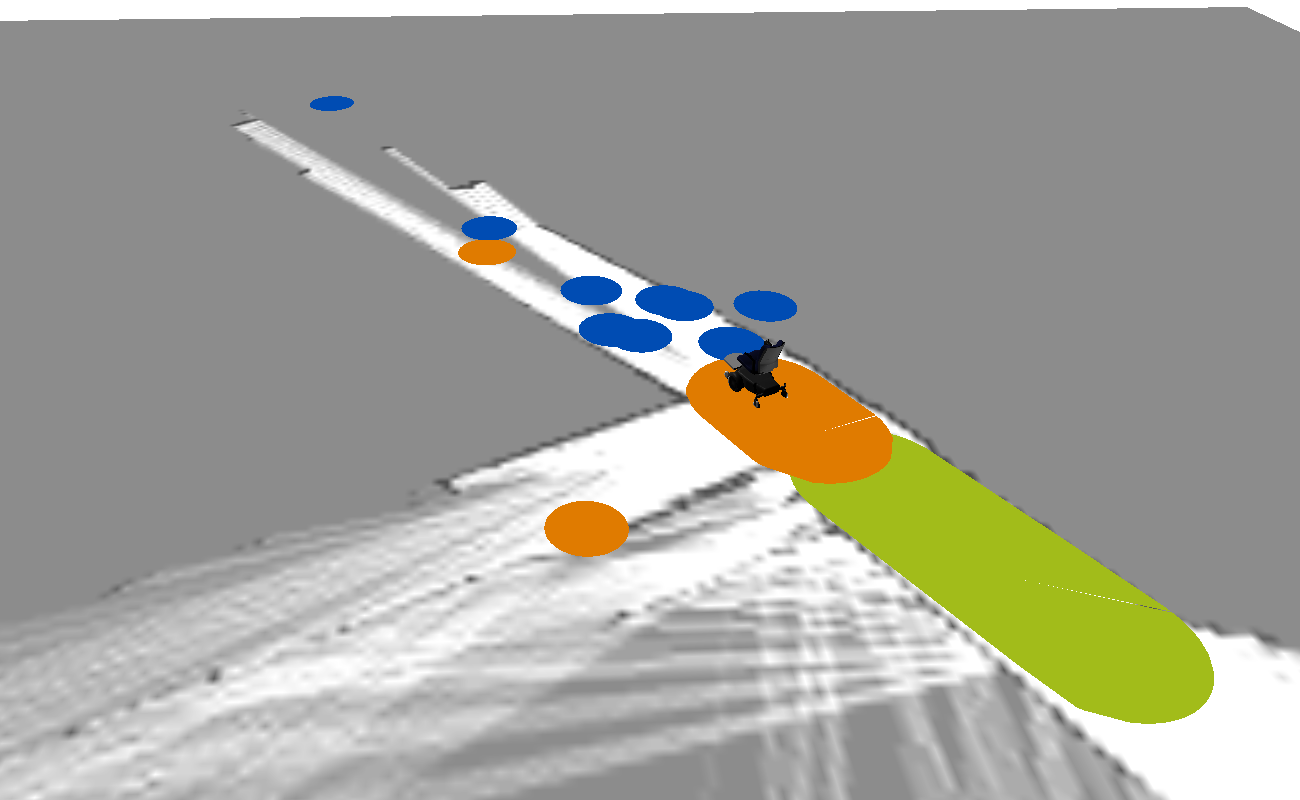}%
  }\hfill%
  \subfloat[\label{fig:value3}$t=8$]{%
    \includeValueGraphic{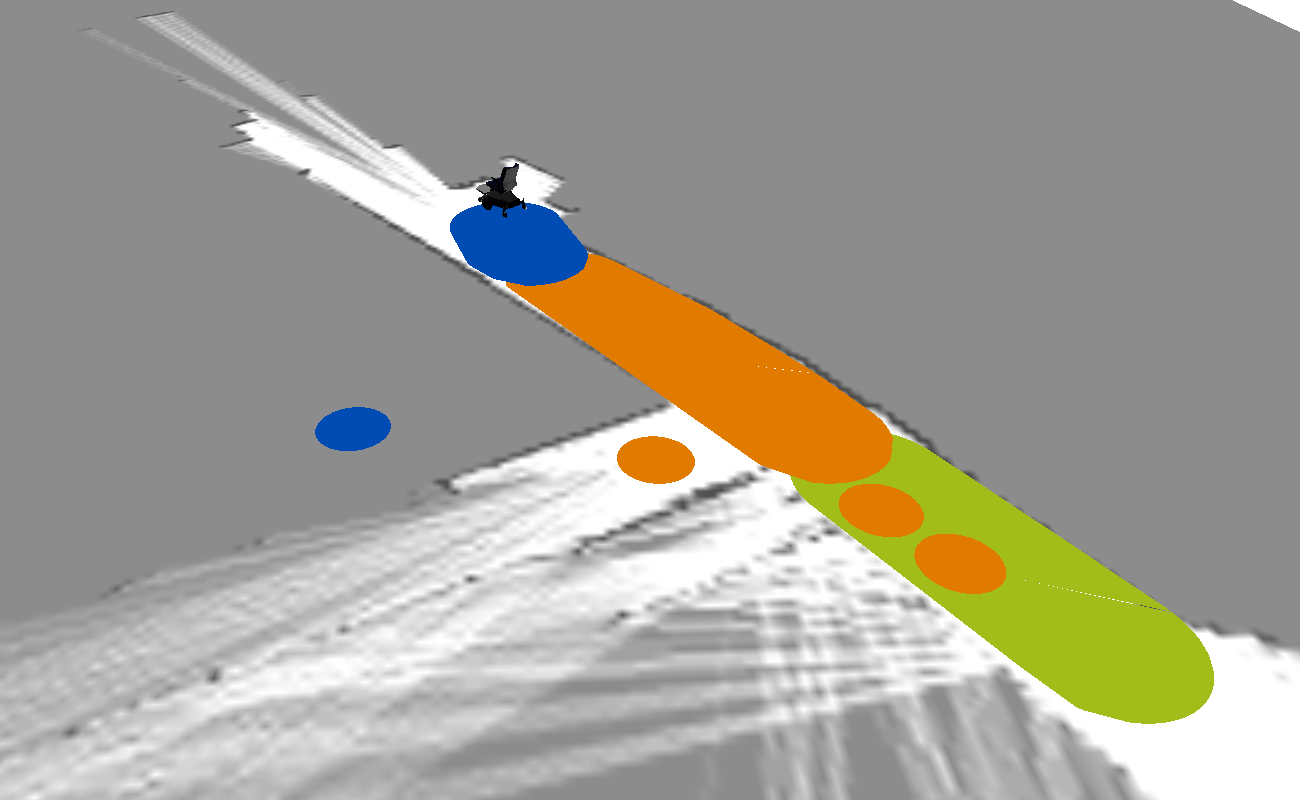}%
  }
  \caption{Visualization of the evolution
    of the semantic map over time as the robot follows the command ``go to the kitchen that
    is down the hallway.''  Small circles and large filled-in areas denote
    sampled and visited regions, respectively, each colored according to
    its type (lab: green, hallway:
    yellow, kitchen: blue). The robot (a) first samples possible locations
    of the kitchen and moves towards them, (b) then observes the hallway
    and refines its estimate using the ``down'' relation provided by the
    user. Finally, the robot (c) reaches the actual kitchen and declares it
    has finished following the direction.}
  \label{fig:belief_world}
\end{figure*}
  % \label{fig:belief_world}

Natural language interpretation becomes particularly challenging when
the expression references areas in the environment unknown to the
robot. Consider an example in which a user directs the voice-commandable
to ``go to the kitchen that is down the
hallway,'' when the wheelchair is in an unknown environment and the
hallway and kitchen are outside the field-of-view of
its sensors (Fig.~\ref{fig:go_to_kitchen_down_hallway}). Unable to
associate the hallway and kitchen with specific locations, most existing
solutions to language understanding would result in the robot exploring until it happens upon
a kitchen. By reasoning over the spatial and
semantic environment information that the command conveys, however, the
robot would be able to follow the spoken directions more efficiently.

In this paper, we propose a framework that follows natural language route
directions within unknown environments by exploiting spatial and semantic
knowledge implicit in the commands. There are three algorithmic
contributions that are integral to our approach. The first is a learned
language understanding model that efficiently infers
environment annotations and desired behaviors from the user's command. The
second is an estimation-theoretic algorithm that learns a distribution over
hypothesized world models by treating the inferred annotations as
observations of the environment and fusing them as observations from the
robot's sensor streams (Fig.~\ref{fig:belief_world}). The third is a belief space policy learned from
human demonstrations that reasons directly over the world model
distribution to identify suitable navigation actions.

This paper generalizes previous work by the authors~\cite{duvallet14},
which was limited to object-relative navigation within small, open
environments. The novel contributions of this work enable robots to follow
natural language route directions in large, complex environments. They
include: a hierarchical framework that learns a compact probabilistic
graphical model for language understanding; a semantic map inference
algorithm that hypothesizes the existence and location of regions in
spatially extended environments; and a belief space policy learned from
human demonstrations that considers spatial relationships with respect to a
hypothesized map distribution.
We demonstrate these advantages through simulations and experiments with a voice-commandable wheelchair in an office-like environment.

% -*- mode:LaTeX; mode:visual-line; mode:flyspell; fill-column:75 -*-

\section{Related Work} \label{sec:related_work}

Recent advancements in language understanding have enabled robots
to understand free-form commands that instruct them to
manipulate objects~\cite{tellex11, howard14a} or navigate through environments using route
directions~\cite{macmahon06, kollar10, chen11, howard14a,matuszek12a}.
With few exceptions, most of these techniques require a
priori knowledge of location, geometry, colloquial name, and type of
all objects and regions within the environment~\cite{kollar10, howard14a,
  tellex11}. Without known world models, however, interpreting free-form
commands becomes much more difficult. Existing methods have dealt with this
by learning a parser that maps the natural language command directly to
plans~\cite{macmahon06, chen11, matuszek12a}. Alternatively, Duvallet et
al.~\cite{duvallet2013} use imitation learning to train a policy that
reasons about uncertainty in the grounding and that is able to backtrack as
necessary. However, none of these approaches explicitly utilize the
knowledge that the instruction conveys to influence their models of the
environment, nor do they reason about its uncertainty. Instead, our
framework treats language as an additional, albeit noisy, sensor that we use
to learn a distribution over hypothesized world models, by taking advantage
of information implicitly contained in a given command.

Related to our algorithm's ability to learn world models, state-of-the-art
semantic mapping frameworks exist that focus on using the robot's sensor
observations to update its representation of the world~\cite{zender08, pronobis10}.
Some methods additionally incorporate natural language descriptions in order to improve the learned world models~\cite{walter13, hemachandra14}.
These techniques, however, only use language to update regions of the environment
that the robot has observed and are not able to extend the maps based on
natural language. Our approach treats natural language as another sensor
and uses it to extend the spatial representation by adding both topological
and metric information regarding hypothesized regions in the environment,
which is then used for planning. Williams et al.~\cite{Williams2013} use a
cognitive architecture to add unvisited locations to a partial
map. However, they only reason about topological relationships to unknown
places, do not maintain multiple hypotheses, and make strong assumptions
about the environment that limit the applicability to real systems.  In
contrast, our approach reasons both topologically and metrically about
regions, and can deal with ambiguity, which allows us to operate in
challenging environments.

% -*- mode:LaTeX; mode:visual-line; mode:flyspell; fill-column:75 -*-

\section{Approach Overview} \label{sec:overview}
We define natural language direction following as one of inferring the
robot's trajectory $x_{t+1:T}$ that is most likely for a given command
$\Lambda^t$:
\begin{equation}
    \argmax{x_{t+1:T} \, \in \, \Re^{n}}
    p\left(x_{t+1:T} | \Lambda^t, z^t, u^t \right),
\label{eqn:problem-statement}
\end{equation}
where $z^t$ and $u^t$ are the history of sensor observations and odometry
data, respectively. Traditionally, this problem has been solved by also
conditioning the distribution over a known world model. Without any a
priori knowledge of the environment, we treat this world model as a latent
variable $S_t$. We then interpret the natural language command in terms of
the latent world model, which results in a distribution over behaviors $\beta_t$. We
then solve the inference problem~\eqref{eqn:problem-statement} by
marginalizing over the latent world model and behaviors:
\begin{equation}
    \begin{split}
        \argmax{x_{t+1:T} \, \in \, \Re^{n}}
        \int\displaylimits_{\beta_t} \int\displaylimits_{S_{t}} p(x_{t+1:T}
        &\vert \beta_t, S_{t}, \Lambda^t) \cdot
        p(\beta_t \vert S_t, \Lambda^t)\\
        &\cdot p(S_{t} \vert \Lambda^t) \, dS_{t} \, d\beta_t,
    \end{split}\label{eqn:marginalization}
\end{equation}
where we have omitted the measurement $z^t$ and odometry $u^t$ histories
for lack of space.

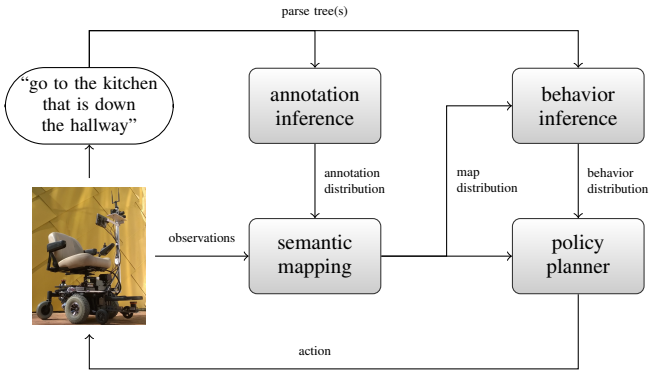
\begin{figure}[!tb]
\begin{center}
\begin{tikzpicture}[textnode/.style={anchor=mid,width=2cm,font=\tiny},
      module/.style={rectangle,rounded corners=3pt,draw=black!80,fill=black!10,minimum size=6mm,font=\footnotesize,top color=white,align=center,text width=1.5cm,minimum width=1.75cm,minimum height=1cm,bottom color=black!20}]
\node[module] (annotation-inference) at (0,0) {annotation inference};
\node[module] (semantic-mapping) at (0,-2) {semantic mapping};
\node[module] (behavior-inference) at (3.5,0) {behavior inference};
\node[module] (policy-planner) at (3.5,-2) {policy planner};
\node[] (robot) at (-3.0,-2.0) {\includegraphics[width=1.5cm]{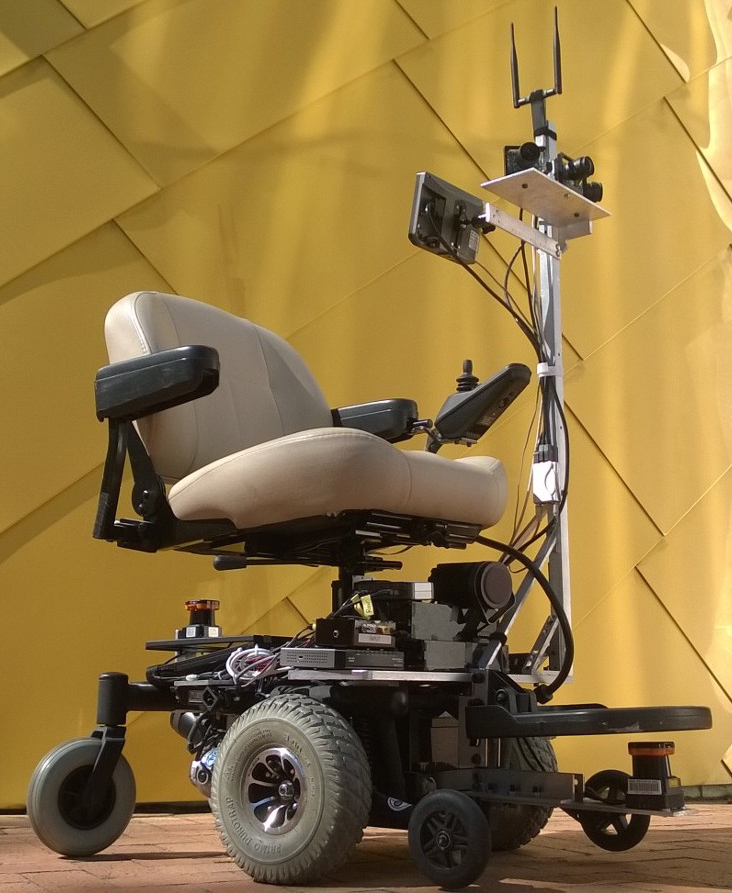}};
\draw[->] (annotation-inference) -- (semantic-mapping);
\draw[->] (semantic-mapping) --(1.75,-2) -- (1.75,0) -- (behavior-inference);
\draw[->] (semantic-mapping) -- (policy-planner);
\draw[->] (behavior-inference) -- (policy-planner);
\node[font=\tiny,align=left,text width=0.875cm] (observationlabel) at (2.325,-1.0) {map distribution};
\node[font=\tiny,align=left,text width=1cm] (annotationdistributionlabel) at (0.625,-1.0) {annotation distribution};
\node[font=\tiny,align=left,text width=1cm] (behaviordistributionlabel) at (4.125,-1.0) {behavior distribution};
\node[font=\scriptsize, draw, rounded corners=15pt, align=center,text width=2cm] (speechlabel) at (-3.0,0) {``go to the kitchen that is down the hallway''};
\draw[->] (robot) -- (speechlabel);
\draw[->] (robot) -- (semantic-mapping);
\draw[->] (speechlabel) -- (-3.0,1.0) -- (0.0,1.0) --  (annotation-inference);
\draw[->] (speechlabel) -- (-3.0,1.0) -- (3.5,1.0) -- (behavior-inference);
\draw[->] (policy-planner) -- (3.5,-3.5) -- (-3.0,-3.5) -- (robot);
\node[font=\tiny,align=center,text width=1cm] (observationslabel) at (-1.5,-1.75) {observations};
\node[font=\tiny,align=center,text width=1cm] (parsetreelabel) at (0,1.25) {parse tree(s)};
\node[font=\tiny,align=center,text width=1cm] (actionlabel) at (0,-3.25) {action};
\end{tikzpicture}
\end{center}
\vspace{-10px}
\caption{Outline of the framework.}
    \label{fig:framework}
\end{figure}

By structuring the problem in this way, we are able to treat inference
as three coupled learning problems. The framework (Fig.~\ref{fig:framework}) first converts
the natural language direction into a set of environment annotations using learned language grounding models. It then treats these
annotations as observations of the environment (i.e., the existence, name,
and relative location of rooms) that it uses together with data from the
robot's onboard sensors to learn a distribution over possible world models
(third factor in Eqn.~\ref{eqn:marginalization}). Our framework then infers
a distribution over behaviors conditioned upon the world model and the
command (second factor). We then solve for the navigation actions that are
consistent with this behavior distribution (first factor) using a learned
belief space policy that commands a single action to the robot.
As the robot executes this action, we update the world
model distribution based upon new utterances and sensor observations, and
subsequently select an updated action according to the policy. This process
repeats as the robot navigates.

The rest of this paper details each of these components in turn.
We then demonstrate our approach to following natural language directions through large unstructured indoor environments on the robot shown in \cref{fig:go_to_kitchen_down_hallway} as well as simulated experiments.
We additionally evaluate our approach to learning belief space policies on a corpus of natural language directions through one floor of an indoor building.

% -*- mode:LaTeX; mode:visual-line; mode:flyspell; fill-column:75 -*-
\section{Natural Language Understanding} \label{sec:nlu}

Our framework relies on learned models to identify the existence of annotations and behaviors conveyed by free-form language and to convert these into a form suitable for semantic mapping and the belief space planner. This is a challenge because of the diversity of natural language directions, annotations, and behaviors.  We perform this translation using the Hierarchical Distributed Correspondence Graph (HDCG) model~\cite{howard14b}, which is a more efficient extension of the Distributed Correspondence Graph (DCG)~\cite{howard14a}. The DCG exploits the grammatical structure of language to formulate a probabilistic graphical model that expresses the correspondence $\phi \in \Phi$ between linguistic elements from the command and their corresponding constituents (\emph{groundings}) $\gamma \in \Gamma$. The factors $f$ in the DCG are represented by log-linear models with feature weights that are learned from a training corpus. The task of grounding a given expression then becomes a problem of inference on the DCG model.

The HDCG model employs DCG models in a hierarchical fashion, by inferring rules $\mathtt{R}$ to construct the space of groundings for lower levels in the hierarchy. At any one level, the algorithm constructs the space of groundings based upon a distribution over the rules from the previous level:
\begin{equation}
    \Gamma \rightarrow \Gamma\left(\mathtt{R}\right).
\end{equation}
The HDCG model treats these rules and, in turn, the structure of the graph, as latent variables. Language understanding then proceeds by performing inference on the marginalized models:
\begin{align}
    &\operatorname*{arg\,max}_\Phi \int_{\mathtt{R}} p\left(\Phi\vert\mathtt{R},\Gamma\left(\mathtt{R}\right),\Lambda,\Psi\right) p\left(\mathtt{R} \vert \Gamma\left(\mathtt{R}\right),\Lambda,\Psi\right)\\
    &\operatorname*{arg\,max}_\Phi \int_{\mathtt{R}} \prod_{i} \prod_{j} f\left(\Phi_{i_{j}},\Gamma_{i_{j}}\left(\mathtt{R}\right),\Lambda_{i},\Psi,\mathtt{R}\right)\times\\
    &\qquad \qquad \qquad \prod_{i} \prod_{j} f\left(\mathtt{R},\Lambda_{i},\Psi,\Gamma_{i_{j}}\left(\mathtt{R}\right)\right).\nonumber
\end{align}

We now describe how the HDCG model infers annotations (representing our knowledge of the environment inferred from the language) and behaviors (representing the intent of the command) to understand the natural language command given by the user.

\subsection{Annotation Inference}

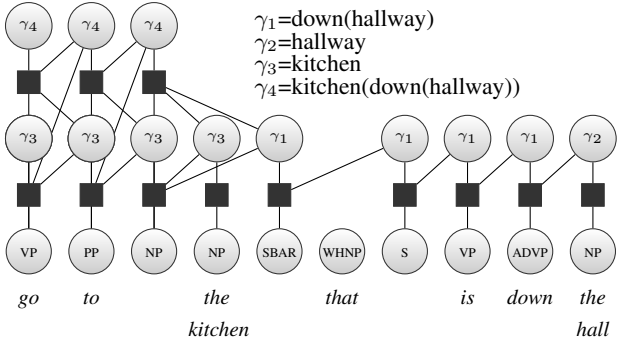
\begin{figure}[tb]
\begin{center}
\begin{tikzpicture}[textnode/.style={anchor=mid,font=\tiny},
      nodeknown/.style={circle,draw=black!80,fill=black!10,minimum size=6mm,font=\tiny,top color=white,bottom color=black!20},
      nodeunknown/.style={circle,draw=black!80,fill=white,minimum size=6mm,font=\tiny},
      factor/.style={rectangle,draw=black!80,fill=black!80,minimum size=3mm,font=\tiny,text=white}]
\node[anchor=mid, align=center](l1) at (0,00.0) {\footnotesize{\textit{go}}};
\node[anchor=mid, align=center](l2) at (0.833,0.0) {\footnotesize{\textit{to}}};
\node[anchor=mid, align=center](l3) at (1.666,0.0) {};
\node[anchor=mid, align=center, text width=0.75cm](l4) at (2.500,0.0) {\footnotesize{\textit{the kitchen}}};
\node[anchor=mid, align=center](l5) at (3.333,0.0) {};
\node[anchor=mid, align=center](l6) at (4.166,0.0) {\footnotesize{\textit{that}}};
\node[anchor=mid, align=center](l7) at (5.000,0.0) {};
\node[anchor=mid, align=center](l8) at (5.833,0.0) {\footnotesize{\textit{is}}};
\node[anchor=mid, align=center](l9) at (6.666,0.0) {\footnotesize{\textit{down}}};
\node[anchor=mid, align=center, text width=0.75cm](l10) at (7.500,0.0) {\footnotesize{\textit{the hall}}};
\node[nodeknown] (l1) at (0.0, 0.625) {};
\node[nodeknown] (l2) at (0.833, 0.625) {};
\node[nodeknown] (l3) at (1.666, 0.625) {};
\node[nodeknown] (l4) at (2.500, 0.625) {};
\node[nodeknown] (l5) at (3.333, 0.625) {};
\node[nodeknown] (l6) at (4.166, 0.625) {};
\node[nodeknown] (l7) at (5.000, 0.625) {};
\node[nodeknown] (l8) at (5.833, 0.625) {};
\node[nodeknown] (l9) at (6.666, 0.625) {};
\node[nodeknown] (l10) at (7.500, 0.625) {};
\node[anchor=mid, align=center](l1label) at (0.0,0.625) {\tiny{VP}};
\node[anchor=mid, align=center](l2label) at (0.833,0.625) {\tiny{PP}};
\node[anchor=mid, align=center](l3label) at (1.666,0.625) {\tiny{NP}};
\node[anchor=mid, align=center](l4label) at (2.500,0.625) {\tiny{NP}};
\node[anchor=mid, align=center](l5label) at (3.333,0.625) {\tiny{SBAR}};
\node[anchor=mid, align=center](l6label) at (4.166,0.625) {\tiny{WHNP}};
\node[anchor=mid, align=center](l7label) at (5.000,0.625) {\tiny{S}};
\node[anchor=mid, align=center](l8label) at (5.833,0.625) {\tiny{VP}};
\node[anchor=mid, align=center](l9label) at (6.666,0.625) {\tiny{ADVP}};
\node[anchor=mid, align=center](l10label) at (7.500,0.625) {\tiny{NP}};
\node[factor] (f12) at (0.0, 2.875) {};
\node[factor] (f22) at (0.833, 2.875) {};
\node[factor] (f32) at (1.666, 2.875) {};
\node[nodeknown] (g12) at (0.0, 3.625) {$\gamma_{4}$};
\node[nodeknown] (g22) at (0.833, 3.625) {$\gamma_{4}$};
\node[nodeknown] (g32) at (1.666, 3.625) {$\gamma_{4}$};
\draw[-] (l1) -- (f12);
\draw[-] (l2) -- (f22);
\draw[-] (l3) -- (f32);
% factors to groundings
\draw[-] (f12) -- (g12);
\draw[-] (f22) -- (g22);
\draw[-] (f32) -- (g32);
\node[factor] (f11) at (0.0, 1.375) {};
\node[factor] (f21) at (0.833, 1.375) {};
\node[factor] (f31) at (1.666, 1.375) {};
\node[factor] (f41) at (2.500, 1.375) {};
\node[factor] (f51) at (3.333, 1.375) {};
\node[factor] (f71) at (5.000, 1.375) {};
\node[factor] (f81) at (5.833, 1.375) {};
\node[factor] (f91) at (6.666, 1.375) {};
\node[factor] (f101) at (7.500, 1.375) {};
\node[nodeknown] (g11) at (0.0, 2.125) {$\gamma_{3}$};
\node[nodeknown] (g21) at (0.833, 2.125) {$\gamma_{3}$};
\node[nodeknown] (g31) at (1.666, 2.125) {$\gamma_{3}$};
\node[nodeknown] (g41) at (2.500, 2.125) {$\gamma_{3}$};
\node[nodeknown] (g51) at (3.333, 2.125) {$\gamma_{1}$};
\node[nodeknown] (g71) at (5.000, 2.125) {$\gamma_{1}$};
\node[nodeknown] (g81) at (5.833, 2.125) {$\gamma_{1}$};
\node[nodeknown] (g91) at (6.666, 2.125) {$\gamma_{1}$};
\node[nodeknown] (g101) at (7.500, 2.125) {$\gamma_{2}$};
% language to factors
\draw[-] (l1) -- (f11);
\draw[-] (l2) -- (f21);
\draw[-] (l3) -- (f31);
\draw[-] (l4) -- (f41);
\draw[-] (l5) -- (f51);
\draw[-] (l7) -- (f71);
\draw[-] (l8) -- (f81);
\draw[-] (l9) -- (f91);
\draw[-] (l10) -- (f101);
% factors to groundings
\draw[-] (f11) -- (g11);
\draw[-] (f21) -- (g21);
\draw[-] (f31) -- (g31);
\draw[-] (f41) -- (g41);
\draw[-] (f51) -- (g51);
\draw[-] (f71) -- (g71);
\draw[-] (f81) -- (g81);
\draw[-] (f91) -- (g91);
\draw[-] (f101) -- (g101);
% factors to other groundings
\draw[-] (f11) -- (g21);
\draw[-] (f11) -- (g22);
\draw[-] (f12) -- (g21);
\draw[-] (f12) -- (g22);
\draw[-] (f21) -- (g31);
\draw[-] (f21) -- (g32);
\draw[-] (f22) -- (g31);
\draw[-] (f22) -- (g32);
\draw[-] (f31) -- (g41);
\draw[-] (f31) -- (g51);
\draw[-] (f32) -- (g41);
\draw[-] (f32) -- (g51);
\draw[-] (f51) -- (g71);
\draw[-] (f71) -- (g81);
\draw[-] (f81) -- (g91);
\draw[-] (f91) -- (g101);
% redraw some
\node[nodeknown] (g11) at (0.0, 2.125) {$\gamma_{3}$};
\node[nodeknown] (g21) at (0.833, 2.125) {$\gamma_{3}$};
\node[nodeknown] (g11) at (0.0, 2.125) {$\gamma_{3}$};
\node[nodeknown] (g21) at (0.833, 2.125) {$\gamma_{3}$};
\node[anchor=mid,align=left,font=\small,text width=2cm] at (4,3.7) {$\gamma_{1}$=down(hallway)};
\node[anchor=mid,align=left,font=\small,text width=2cm] at (4,3.4) {$\gamma_{2}$=hallway};
\node[anchor=mid,align=left,font=\small,text width=2cm] at (4,3.1) {$\gamma_{3}$=kitchen};
\node[anchor=mid,align=left,font=\small,text width=2cm] at (4,2.8) {$\gamma_{4}$=kitchen(down(hallway))};
\end{tikzpicture}
\end{center}
\vspace{-10px}
\caption{The active groundings in annotation inference for the direction ``go to the kitchen that is down the hall''.  The two symbols at the root of the sentence ($\gamma_{3}$,$\gamma_{4}$) are sent to the semantic map to fuse with other observations.}
\label{fig:annotation-inference}
\end{figure}

An annotation is a set of object types and subspaces.  A subspace is defined here as a spatial relationship (e.g., down, left, right) with respect to an object type.  In the experiments described in Section \ref{sec:results} we assume 17 object types and 12 spatial relationships.  We also permit object types to express a spatial relationship with another object type.  We denote object types by their physical type (e.g., kitchen, hallway), subspaces as the relationship type with an object type argument (e.g., down(kitchen), left(hallway)), and object types with spatial relationships as an object type with a subspace argument (e.g., kitchen(down(hallway))).  Since the number of possible combinations of annotations is equal to the power set of the number of symbols, $2^{3,485}$ annotations can be expressed by an instruction.\footnote{3,485 symbols = 17 object types, 204 subspaces, and 3,264 object types with spatial relationships (we exclude object types with spatial relationships to the same object type)}  The HDCG model infers a distribution of graphical models to efficiently generate annotations by assuming conditional independence of constituents and eliminating symbols that are learned to be irrelevant to the utterance.  For example, Figure \ref{fig:annotation-inference} illustrates the model for the direction ``go to the kitchen that is down the hall.''  In this example only 4 of the 3,485 symbols (two object types, one subspace, and one object type with a spatial relationship) are active in this model.  Note that all factors with inactive correspondence variables are not illustrated in Figures \ref{fig:annotation-inference} and \ref{fig:behavior-inference}.  At the root of the sentence the symbols for an object type (kitchen) and an object type with a spatial relationship (kitchen(down(hallway))) are sent to the semantic map to fuse with other observations.

\subsection{Behavior Inference}
\begin{figure}[tb]
\begin{center}
\begin{tikzpicture}[textnode/.style={anchor=mid,font=\tiny},
      nodeknown/.style={circle,draw=black!80,fill=black!10,minimum size=6mm,font=\tiny,top color=white,bottom color=black!20},
      nodeunknown/.style={circle,draw=black!80,fill=white,minimum size=6mm,font=\tiny},
      factor/.style={rectangle,draw=black!80,fill=black!80,minimum size=3mm,font=\tiny,text=white}]
\node[anchor=mid, align=center](l1) at (0,00.0) {\footnotesize{\textit{go}}};
\node[anchor=mid, align=center](l2) at (0.833,0.0) {\footnotesize{\textit{to}}};
\node[anchor=mid, align=center](l3) at (1.666,0.0) {};
\node[anchor=mid, align=center, text width=0.75cm](l4) at (2.500,0.0) {\footnotesize{\textit{the kitchen}}};
\node[anchor=mid, align=center](l5) at (3.333,0.0) {};
\node[anchor=mid, align=center](l6) at (4.166,0.0) {\footnotesize{\textit{that}}};
\node[anchor=mid, align=center](l7) at (5.000,0.0) {};
\node[anchor=mid, align=center](l8) at (5.833,0.0) {\footnotesize{\textit{is}}};
\node[anchor=mid, align=center](l9) at (6.666,0.0) {\footnotesize{\textit{down}}};
\node[anchor=mid, align=center, text width=0.75cm](l10) at (7.500,0.0) {\footnotesize{\textit{the hall}}};
\node[nodeknown] (l1) at (0.0, 0.625) {};
\node[nodeknown] (l2) at (0.833, 0.625) {};
\node[nodeknown] (l3) at (1.666, 0.625) {};
\node[nodeknown] (l4) at (2.500, 0.625) {};
\node[nodeknown] (l5) at (3.333, 0.625) {};
\node[nodeknown] (l6) at (4.166, 0.625) {};
\node[nodeknown] (l7) at (5.000, 0.625) {};
\node[nodeknown] (l8) at (5.833, 0.625) {};
\node[nodeknown] (l9) at (6.666, 0.625) {};
\node[nodeknown] (l10) at (7.500, 0.625) {};
\node[anchor=mid, align=center](l1label) at (0.0,0.625) {\tiny{VP}};
\node[anchor=mid, align=center](l2label) at (0.833,0.625) {\tiny{PP}};
\node[anchor=mid, align=center](l3label) at (1.666,0.625) {\tiny{NP}};
\node[anchor=mid, align=center](l4label) at (2.500,0.625) {\tiny{NP}};
\node[anchor=mid, align=center](l5label) at (3.333,0.625) {\tiny{SBAR}};
\node[anchor=mid, align=center](l6label) at (4.166,0.625) {\tiny{WHNP}};
\node[anchor=mid, align=center](l7label) at (5.000,0.625) {\tiny{S}};
\node[anchor=mid, align=center](l8label) at (5.833,0.625) {\tiny{VP}};
\node[anchor=mid, align=center](l9label) at (6.666,0.625) {\tiny{ADVP}};
\node[anchor=mid, align=center](l10label) at (7.500,0.625) {\tiny{NP}};
\node[factor] (f11) at (0.0, 1.375) {};
\node[factor] (f21) at (0.833, 1.375) {};
\node[factor] (f31) at (1.666, 1.375) {};
\node[factor] (f41) at (2.500, 1.375) {};
\node[factor] (f51) at (3.333, 1.375) {};
\node[factor] (f71) at (5.000, 1.375) {};
\node[factor] (f81) at (5.833, 1.375) {};
\node[factor] (f91) at (6.666, 1.375) {};
\node[factor] (f101) at (7.500, 1.375) {};
\node[nodeknown] (g11) at (0.0, 2.125) {$\gamma_{8}$};
\node[nodeknown] (g21) at (0.833, 2.125) {$\gamma_{7}$};
\node[nodeknown] (g31) at (1.666, 2.125) {$\gamma_{7}$};
\node[nodeknown] (g41) at (2.500, 2.125) {$\gamma_{7}$};
\node[nodeknown] (g51) at (3.333, 2.125) {$\gamma_{5}$};
\node[nodeknown] (g71) at (5.000, 2.125) {$\gamma_{5}$};
\node[nodeknown] (g81) at (5.833, 2.125) {$\gamma_{5}$};
\node[nodeknown] (g91) at (6.666, 2.125) {$\gamma_{5}$};
\node[nodeknown] (g101) at (7.500, 2.125) {$\gamma_{6}$};
% language to factors
\draw[-] (l1) -- (f11);
\draw[-] (l2) -- (f21);
\draw[-] (l3) -- (f31);
\draw[-] (l4) -- (f41);
\draw[-] (l5) -- (f51);
\draw[-] (l7) -- (f71);
\draw[-] (l8) -- (f81);
\draw[-] (l9) -- (f91);
\draw[-] (l10) -- (f101);
% factors to groundings
\draw[-] (f11) -- (g11);
\draw[-] (f21) -- (g21);
\draw[-] (f31) -- (g31);
\draw[-] (f41) -- (g41);
\draw[-] (f51) -- (g51);
\draw[-] (f71) -- (g71);
\draw[-] (f81) -- (g81);
\draw[-] (f91) -- (g91);
\draw[-] (f101) -- (g101);
% factors to other groundings
\draw[-] (f11) -- (g21);
\draw[-] (f21) -- (g31);
\draw[-] (f31) -- (g41);
\draw[-] (f31) -- (g51);
\draw[-] (f51) -- (g71);
\draw[-] (f71) -- (g81);
\draw[-] (f81) -- (g91);
\draw[-] (f91) -- (g101);
% redraw some
\node[nodeknown] (g11) at (0.0, 2.125) {$\gamma_{8}$};
\node[nodeknown] (g21) at (0.833, 2.125) {$\gamma_{7}$};
\node[] (image) at (5,4.5) {\includegraphics[width=0.225\textwidth]{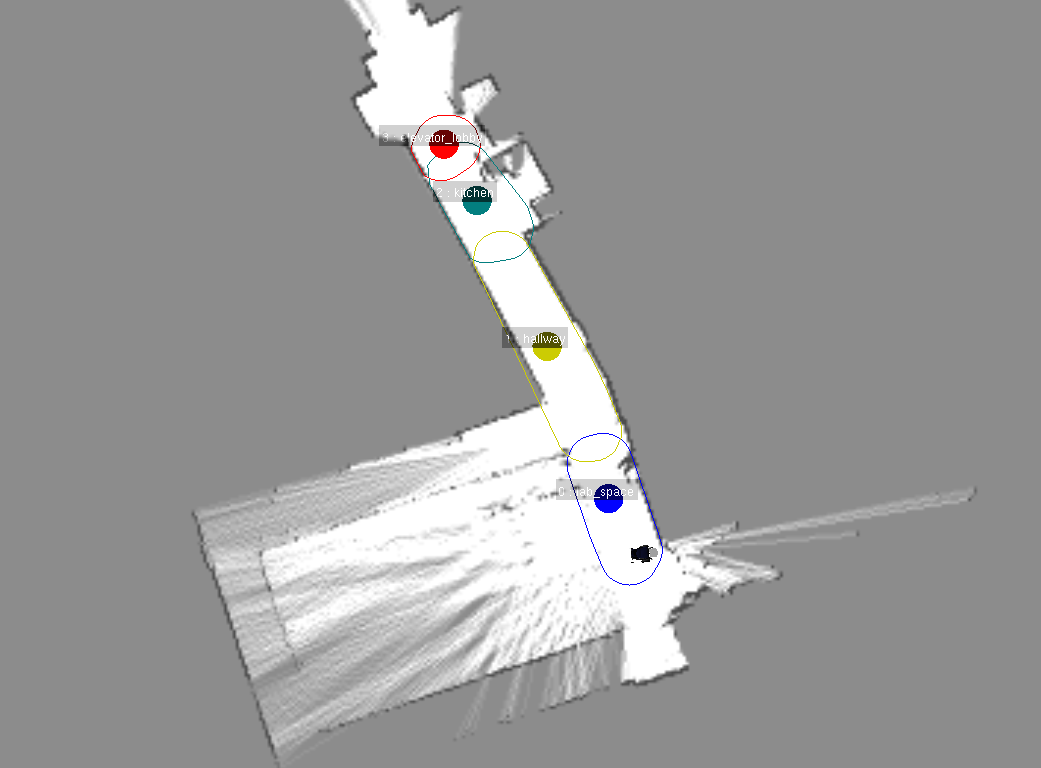}};
\node[anchor=mid,align=left,font=\tiny,text width=2cm] (o1) at (5.9,5.2) {$o_{1}$(kitchen)};
\node[anchor=mid,align=left,font=\tiny,text width=2cm] (o2) at (5.7,5.6) {$o_{3}$(elevator lobby)};
\node[anchor=mid,align=left,font=\tiny,text width=2cm] (o3) at (6.2,4.65) {$o_{2}$(hallway)};
\node[anchor=mid,align=left,font=\tiny,text width=2cm] (o4) at (6.4,4.2) {$o_{4}$(lab space)};
\node[anchor=mid,align=left,font=\tiny,text width=2cm] (robot) at (6.5,3.7) {robot};
%0
\node[anchor=mid,align=left,font=\normalsize,text width=2cm] at (0.7,5.25) {$\gamma_{5}$=down($o_{2}$)};
\node[anchor=mid,align=left,font=\normalsize,text width=2cm] at (0.7,4.75) {$\gamma_{6}$=$o_{2}$};
\node[anchor=mid,align=left,font=\normalsize,text width=2cm] at (0.7,4.25) {$\gamma_{7}$=$o_{1}$};
\node[anchor=mid,align=left,font=\normalsize,text width=2cm] at (0.7,3.75) {$\gamma_{8}$=navigate($o_{1}$)};
\end{tikzpicture}
\end{center}
\vspace{-10px}
\caption{The active groundings in behavior inference for the direction ``go to the kitchen that is down the hall'' in the context of a inferred map with 4 objects.  In this example a \textit{navigate} action with a goal relative to $o_{1}$ would be sent to the policy planner.  }
\label{fig:behavior-inference}
\end{figure}

A behavior is a set of objects, subspaces, actions, objectives, and constraints.  Behavior inference differs from annotation inference by considering objects from the semantic map and subspaces defined with respect to objects from the semantic map instead of only object types.  We denote actions by their  type and an object or subspace argument (e.g., navigate(hallway)), objectives by their type (e.g., quickly, safely), and constraints as objects with spatial relationship from the semantic map (e.g., $o_{4}$(down($o_{3}$))).  In the experiments presented in Section \ref{sec:results} we assume 4 action types, 3 objectives, and 12 spatial relations.  Just as with annotation inference, the HDCG model eliminates irrelevant action types, objective types, objects, and spatial relationships to efficiently infer behaviors.  Figure \ref{fig:behavior-inference} illustrates the model for the direction ``go to the kitchen that is down the hall'' in the context of an inferred map.  In this example a \textit{navigate} action with a goal relative to $o_{1}$ would be inferred as the most likely behavior for the policy planner.

% -*- mode:LaTeX; mode:visual-line; mode:flyspell; fill-column:75 -*-

\section{Semantic Mapping}

We represent the world model as a modified \emph{semantic
  map}~\cite{walter13} $S_t = \{G_t, X_t\}$, a hybrid metric and
topological representation of the environment. The topology $G_t$
consists of nodes $n_i$ that denote locations in the environment, edges
that denote
inter-node connections, and non-overlapping
regions $R_\alpha = \{n_1, n_2, \ldots, n_m\}$ that represent spatially
coherent areas compatible with a human's
decomposition of space (e.g., rooms and hallways).  We associate a pose
$x_i$ with each node $n_i$, the vector of which constitutes the metric map
$X_t$. Each region is also labeled according to its type
(e.g., kitchen, hallway). An edge connects two regions that the robot has
transitioned between or for which language indicates the
existence of an inter-region spatial relation (e.g., that the kitchen is
``down'' the hallway).

Annotations extracted from a given command provide information regarding
the existence, relative location, and type of regions\footnote{Regions as
  defined by the mapping framework are also considered as objects for the
  purpose of natural language understanding.} in the environment. We learn a
distribution over world models consistent with these annotations by
treating them as observations $\alpha_t$ in a filtering framework. We
combine these observations with those from other sensors onboard the robot
(LIDAR and region appearance observations) $z_t$ to maintain a distribution
over the semantic map:
\begin{subequations}
  \begin{align}
    p(S_t \vert \Lambda^t, z^t, u^t)\! &\approx p(S_t \vert
    \alpha^t, z^t, u^t)\\
    \!&= p(G_t, X_t, \vert \alpha^t, z^t, u^t)\\
    \!&= p(X_t \vert G_t, \alpha^t, z^t, u^t) p(G_t \vert \alpha^t, z^t, u^t),
  \end{align}
\end{subequations}
where we assume that an utterance $\Lambda^t$ provides a set of annotations
$\alpha_t$. The factorization within the last line models the metric map
induced by the topology, as with pose graph representations~\cite{kaess08}.
We maintain this distribution over time using a Rao-Blackwellized particle
filter (RBPF)~\cite{doucet00}, with a sample-based approximation of the
distribution over the topology, and a Gaussian distribution over metric
poses.

The robot observes transitions between environment regions and the semantic
label of its current region. As scene understanding is not the focus of
this work, we use AprilTag fiducials~\cite{Olson2011} placed in each region
that denotes its label. Unlike our earlier work~\cite{hemachandra14} in
which we segment regions based only on their spatial coherence using
spatial clustering, here we additionally use the presence of conflicting
spatial appearance tags to also segment the region. As such, we assume that
we are aware of the segmentation of the space immediately, which is not
possible with a purely spectral clustering based approach, allowing us to
immediately evaluate each particle's likelihood based on the observation of
region appearance. In turn, we can down-weight particles that are
inconsistent with the actual layout of the world sooner, reducing the
number of actions the robot must take to satisfy the command.

We maintain each particle through the three steps of the RPBF. First, we
propagate the topology by sampling modifications to the graph when the
robot receives new sensor observations or annotations. Second, we perform a
Bayesian update to the pose distribution based upon the sampled
modifications to the underlying graph. Third, we update the weight of each
particle based on the likelihood of generating the given observations, and
resample as needed to avoid particle depletion. We now outline this process
in more detail.

%% \todomw{This could go in the results section.}
%% \todoinsachi{This following section is necessary here - to outline that we
%%   segment immediately - but seems to break the flow between the above and
%%   below sections}

%% \todoinsachi{The robot makes two observations from its onboard sensors; it
%%   observes when it transitions from one region to another, and it observes
%%   the label of its current region. In this implementation, we enable this
%%   by placing AprilTag fiducials~\cite{Olson2011} in each region. }

%%\todomw{This paragraph seems out of place, given previous paragraph.}

%% particles with sampled regions of different region types sooner, reducing
%% the amount of actions we are required to take to satisfy the command.

During the proposal step, we first add an additional node~$n_t$ and edge to
each particle's topology that model the robot's motion $u_t$, yielding a new
topology $S_t^{(i)-}$. We then sample modifications to the topology
$\Delta_t^{(i)} = \{\Delta_{\alpha_t}^{(i)}, \Delta_{z_t}^{(i)}\}$ based on
the most recent annotations $\alpha_t$ and sensor observations $z_t$:
%\begin{equation}
\begin{multline}\label{eq:sample_graph}
    p(S_t^{(i)} | S_{t-1}^{(i)}, \alpha_t, z_t, u_t)=p(\Delta_{\alpha_t}^{(i)} | S_{t}^{(i)-}, \alpha_t)\, \\
    p(\Delta_{z_t}^{(i)} | S_{t}^{(i)-}, z_t)\,
    p(S_t^{(i)-} | S_{t-1}^{(i)}, u_t).
\end{multline}
This updates the proposed graph topology $S_{t}^{(i)-}$ with the graph
modifications $\Delta_t^{(i)}$ to yield the new semantic map
$S_t^{(i)}$. The updates can include the addition and deletion of nodes
and regions from the graph that represent newly hypothesized or observed
regions, and edges that express
express spatial relations inferred from observations or annotations.

%% During the proposal step, we first augment each sample topology with an
%% additional node and edge that model the robot's motion $u_t$, resulting in
%% a new topology $S_t^{(i)-}$. We then sample modifications to the graph
%% $\Delta_t^{(i)} = \{\Delta_{\alpha_t}^{(i)}, \Delta_{z_t}^{(i)}\}$ based
%% upon the most recent annotations ($\alpha_t$) and sensor observations
%% ($z_t$):
%% %
%% %\begin{equation}
%% \begin{multline}\label{eq:sample_graph}
%%     p(S_t^{(i)} | S_{t-1}^{(i)}, \alpha_t, z_t, u_t)=p(\Delta_{\alpha_t}^{(i)} | S_{t}^{(i)-}, \alpha_t)\, \\
%%     p(\Delta_{z_t}^{(i)} | S_{t}^{(i)-}, z_t)\,
%%     p(S_t^{(i)-} | S_{t-1}^{(i)}, u_t).
%% \end{multline}
%% This updates the proposed graph topology $S_{t}^{(i)-}$ with the graph modifications $\Delta_t^{(i)}$
%% to yield the new semantic map $S_t^{(i)}$. The updates can include the addition
%% of nodes and regions to the graph representing newly hypothesized or observed
%% objects. They also may include the addition of edges between nodes to
%% express spatial relations inferred from observations or annotations.

We sample graph modifications from two independent proposal
distributions for annotations $\alpha_t$ and robot observations $z_t$.
This is done by sampling a grounding for each observation and modifying the
graph according to the implied grounding.
%% The process samples a grounding for each observation, which in turn
%% determines the modifications to the graph. \todomw{This last sentence is vague.}

\subsection{Graph modifications based on natural language}
%% \todoinsachi{language sampling feels like it belongs before the spatial
%%   stuff - because we end up talking about grounding sampled constraints to
%%   new spaces}

Given a set of annotations \mbox{$\alpha_t=\{\alpha_{t,j}\}$}, we sample
modifications to the graph for each particle.
%
%
%% \begin{equation}
%%   p(\Delta_{\alpha_t}^{(i)} | S_{t}^{(i)-}, \alpha_t)  = \prod_j
%%   p(\Delta_{\alpha_{t,j}}^{(i)} | S_{t}^{(i)-}, \alpha_{t,j}). \label{eq:proposal_annotation}
%% \end{equation}
%
%
An annotation $\alpha_{t,j}$ contains a spatial relation and figure when
the language describes one region (e.g., ``go to the elevator lobby''), and
an additional landmark when the language describes the relation between two
regions (e.g., ``go to the lobby through the hallway''). We use a
likelihood model over the spatial relation to sample landmark and figure
pairs for the grounding. This model employs a Dirichlet process prior that
accounts for the fact that the annotation may refer to regions that exist
in the map or to unknown regions. If either the landmark or the figure are
sampled as new regions, we add them to the graph and create an edge
between them. We also sample the metric constraint associated with this
edge based on the spatial relation. The spatial relation models employ
features that describe the locations of the regions, their boundaries,
and robot's location at the time of the utterance, and are trained
based upon a natural language corpus~\cite{tellex11}.

%% \subsubsection{Sampling edges based on natural language}\todo{Can get rid of the description of the sampling process}
%% When only one of the landmark or the figure is grounded to a visited region in
%% the map, we sample the constraint to the remaining spatial entity by
%% randomly sampling a set of potential locations in the environment and
%% evaluating the likelihood of these locations conforming with the spatial
%% relation, and selecting the maximum likelihood sampled constraint.

%% When language only provides information about the presence of a location
%% (e.g., ``go to the kitchen''), or when both entities are ungrounded, we
%% sample the landmark based on a spatially aware prior, such that new regions
%% are not sampled on top of existing regions (by making use of the current
%% locations of frontier regions).

\subsection{Graph modifications based on robot observations}

%% \todoinsachi{The robot makes two types of observations from its onboard
%%   sensors; it detects when it transitions from one region to another, and
%%   it observes the label of its current region.}

%%Pulled from the thesis
If the robot does not observe a region transition (i.e. the robot is in the
same region as before), the algorithm adds the new node~$n_t$ to the
current region and modifies its spatial extent. If there are any edges
denoting spatial relations to hypothesized regions, the algorithm resamples
their constraint if its likelihood changes significantly due to the
modified spatial extent of the current region.
%% This is done
%% because some of the features that are used to calculate the likelihood of a
%% constraint are dependent on the region's spatial extent, which grows as the
%% robot observes more of the region.

Alternatively, if the robot observes a region transition, the new
node~$n_t$ is assigned to a new or existing region as follows. First, the
algorithm checks if the robot is in a previously visited region, based on
spatial proximity, in which case it will add $n_t$ to that
region. Otherwise, it will create a new region and check whether it
matches a region that was previously hypothesized based on an annotation
(for example, a newly-visited kitchen can be the same as a hypothesized
kitchen described with language).
We do so by sampling a grounding to any unobserved regions in the topology
using a Dirichlet process prior.
If this process results in a grounding to an existing hypothesized region,
we remove the hypothesized region and adjust the topology accordingly,
resampling any edges to yet-unobserved regions.  For example, if an
annotation suggested the existence of a ``kitchen down the hallway,'' and
we grounded the robot's current region to the hypothesized hallway, we
would reevaluate the ``down'' relation for the hypothesized kitchen with
respect to this detected hallway.

\subsection{Re-weighting particles and resampling}

After modifying each particle's topology, we perform a Bayesian
update to its Gaussian distribution. We then re-weight each particle
according to the likelihood of generating language annotations and
region appearance observations:
\begin{equation}\label{eq:weight_update_all}
    w_t^{(i)}\!\!=\! p(z_t, \alpha_t | S_{t-1}^{(i)}) w_{t-1}^{(i)} \!\!=\! p(\alpha_t| S_{t-1}^{(i)}) p(z_t| S_{t-1}^{(i)}) w_{t-1}^{(i)}.
\end{equation}
%
%
%\todoinsachi{This is especially needed to account for negative observations}

%% This step accounts for negative observations that are not considered in the
%% proposal model.
When calculating the likelihood of each region appearance observation, we
consider the current node's region type and calculate the likelihood of
generating this observation given the topology. %% given the unobserved regions in the particle.
In effect, this down-weights any particle with a sampled
region of a particular type existing on top of a known traversed region of
a different type.  We use a likelihood model that describes the observation
of a region's type, with a latent binary variable $v$ that denotes whether or not
the observation is valid.
%% Our observation models the likelihood of receiving a valid region
%% appearance observing a particular region
%% appearance, given the robot's location w.r.t. the region, and the
%% likelihood of the
%% In order
%% to do this, we introduce a latent variable $v$ that denotes the validity of the
%% sampled region generating a given observation, model the likelihood of
%% generating a particular region appearance observation, given the region can
%% generate a valid region observation at the robot's current
%% location. \todo{This sentence doesn't make sense}
We marginalize over $v$ to arrive at the likelihood of generating the given
observation, where $R_u$ is the set of unobserved regions in particle
$S_{t-1}^{(i)}$:
\begin{equation}\label{eq:likelihood_appearance}
  p(z^t\vert{S_{t-1}^{(i)}}) = \prod_{R_i\in {R_u}}\left(\sum\limits_{v\in {1,0}}{p(z^t\vert{v,R_i}) \times p(v\vert{R_i})}\right).
\end{equation}
For annotations, we use the language grounding likelihood under the
map at the previous time step. As such, a particle with an existing pair of
regions conforming to a specified language constraint will be weighted
higher than one without.
When the particle weights fall below a threshold, we resample particles to
avoid particle depletion~\cite{doucet00}.

\section{Reasoning and Learning in Belief Space} \label{sec:planning}

Searching for the complete trajectory that is optimal in the distribution of maps would be intractable.
Instead, we treat direction following as sequential decision making under uncertainty, where a policy $\pi$ minimizes a single step of the cost function $c$ over the available actions $a \in A_t$ from state~$x$:
\begin{equation} \label{eqnPolicy}
  \pi \parens{x, \Mapt} = \argmin{a \in A_t} c\parens{x, a, \Mapt}.
\end{equation}
After executing the action and updating the map distribution, we repeat this process until the policy declares it has completed following the direction using a separate stop action.

As the robot travels in the environment, it keeps track of the nodes in the topological graph $G_t$ it has visited ($\visited$) and frontiers ($\frontiers$) that lie at the edge of explored space.
The action set $A_t$ consists of paths to nodes in the graph.
An additional action $\astop$ declares that the policy has completed following the direction.
Intuitively, an action represents a single step along the path that takes the robot towards its destination. Each action may explore new parts of the environment (for example continuing to travel down a hallway) or backtrack if the policy has made a mistake (for example, traveling to a room in a different part of the environment).
The following sections explain how the policy reasons in belief space, and the novel imitation learning formulation to train the policy from demonstrations of correct behavior.  % Didn't fit.

\subsection{Belief Space Reasoning using Distribution Embedding}
The semantic map $\Mapt$ provides a distribution over the possible locations of the landmarks relevant to the command the robot is following.
As such, the policy $\pi$ must reason about a \emph{distribution} of action features when computing the cost of any action $a$.
We accomplish this by embedding the action feature distribution in a Reproducing Kernel Hilbert Space (RKHS), using the mean feature map~\cite{smola2007} consisting of the first $K$ moments of the features computed with respect to each map sample $\Mapti$ (and its likelihood):
\begin{align} \label{eqnPolicyDecomposition}
  \Momenta \parens{x, a, \Mapt} & = \sum_{\Mapti} p(\Mapti) \: \phiXAMapi \\
  \Momentb \parens{x, a, \Mapt} & = \sum_{\Mapti} p(\Mapti) \: \parens{\phiXAMapi - \Momenta }^2\\
  & \ldots \nonumber \\
  \Momentk \parens{x, a, \Mapt} & = \sum_{\Mapti} p(\Mapti) \: \parens{\phiXAMapi - \Momenta }^k
\end{align}
Intuitively, this formulation computes features for the action and all hypothesized landmarks individually, aggregates these feature vectors, and then computes moments of the feature vector distribution (mean, variance, and higher order statistics).
A simplified illustration, shown in \autoref{figFeatureComputation}, shows how our approach computes belief space features for two actions with a hypothesized kitchen (with two possible locations).

The cost function in \autoref{eqnPolicy} can now be rewritten as a weighted sum of the first $K$ moments of the feature distribution:
\begin{equation} \label{eqnBeliefPolicySumK}
  \costMapt = \sum_{i=1}^K \witranspose \; \Momenti \XAMapt.
\end{equation}
By concatenating the weights and moments into respective column vectors $W := [w_1; \ldots; w_k]$ and $F := [\Momenta; \ldots; \Momentk]$, we can rewrite the policy in \autoref{eqnPolicy} as minimizing a weighted sum of the feature moments $F_a$ for action $a$ :
\begin{equation} \label{eqnBeliefPolicyVectorK}
  \pi \parens{x, \Mapt} = \argmin{a \in A_t} W^T F_a.
  %\costMapt = W^T F_a.
\end{equation}

% NOTE: W \in \R^{KN}

% -*- mode:LaTeX; mode:visual-line -*-

% -*- mode:LaTeX; mode:visual-line -*-

% Sample direction: Turn right when you see the _____
% Take a right to the ___________

\usetikzlibrary{positioning}

\newcommand{\robotX}{-3.5}
\newcommand{\robotY}{+.2}
\newcommand{\wallLengthBeyondRobot}{2.0}
\newcommand{\lowerWall}{-0.7}

\newcommand{\kitchenAX}{-8.0}
\newcommand{\kitchenAY}{1.5}
\newcommand{\kitchenBX}{-8.75}
\newcommand{\kitchenBY}{0.0}

\tikzstyle{invisibleLandmark}=[
  draw=black, fill=black!05, thick, dashed]

\tikzstyle{kitchenStyle}=[
  rectangle,
  draw,
  dashed,
  fill=black!15,
  minimum width=4em,
  minimum height=1.5em,
  label={center:Kitchen}
]

\newcommand{\drawBeliefSpace}{
  %\drawGrid

  \node[] (startPosition) at (0, 0.2) {};
  \node[] (robotPosition) at (\robotX, \robotY) {};
  \node[] (actionStraight) at (\robotX-\wallLengthBeyondRobot, .2) {};
  \node[] (midKitchen) at (\robotX, .2) {};
  \node[] (actionRight) at (\robotX, 2) {};

  % Draw an ellipse around the kitchens.
  \node[] (centroid) at (-8.5,0.9){};
  \draw[rotate around={-55:(centroid.center)},
    fill=black!05,
    dashed,
  ] (centroid.center) ellipse (1.7cm and 2cm);

  \draw[visibleWallStyle]  % Wall on top, right, and bottom
  (\robotX+1, 2) -- (\robotX+1, 1)  -- (1, 1) --  (1, \lowerWall) --
  (\robotX-\wallLengthBeyondRobot, \lowerWall);

  \draw[visibleWallStyle]  % Wall around left corner.
  (\robotX-1, 2) -- (\robotX-1, 1)  -- (\robotX-\wallLengthBeyondRobot, 1) ;

  \def\shift{0.5mm}  % Action to the right.
  \draw[policyActionStyle, transform canvas={yshift=\shift}, shorten >=\shift]
  (startPosition.west) -- (midKitchen.center) -- (actionRight.center);

  \def\shift{-0.5mm}  % Action to the left.
  \draw[policyActionStyle, transform canvas={yshift=\shift}, shorten >=\shift]
  (startPosition.west) -- (actionStraight.center);

  % Put frontier nodes.
  \node[frontierNode, label=right:$a_1$] (frontierKitchen) at (actionRight) {};
  \node[frontierNode, label=below:$a_2$] (frontierStair) at (actionStraight) {};

  \node (kitchenA) at (\kitchenAX,\kitchenAY) [kitchenStyle] {};
  \node (kitchenB) at (\kitchenBX,\kitchenBY) [kitchenStyle] {};

  \drawRobot{robotPosition}
  \drawStart
}

\tikzstyle{features1Style}=[
  thick,
  dotted,
  rounded corners,
  color=blue,
]

\tikzstyle{features2Style}=[
  thick,
  dotted,
  rounded corners,
  color=red,
]

\newcommand{\drawFeatureArrows}{
  \draw[features1Style] (actionRight) -- (kitchenA.east);
  \draw[features1Style] (actionRight) -- (kitchenB.east);
  \node[features1Style] at (\robotX-1.5, 2.75) {$\phi(a_1, S^1), \phi(a_1, S^2)$};

  \draw[features2Style] (actionStraight) -- (kitchenA.east);
  \draw[features2Style] (actionStraight) -- (kitchenB.east);
  \node[features2Style] at (\robotX-3, -1.2) {$\phi(a_2, S^1), \phi(a_2, S^2)$};
}

\tikzstyle{visibleLandmark}=[
  draw=black, fill=black!20, thick,
  font = \footnotesize,
]

\tikzstyle{landmarkText} = [
  text centered,
  font = \footnotesize]

\newcommand{\nodeSize}{7}
\tikzstyle{visibleNode}=[
  circle,
  draw=black!80,
  fill=blue!10,
  thick,
  inner sep = 0,
  minimum size = \nodeSize]
\tikzstyle{frontierNode}=[
  circle, double,
  draw=black!80,
  fill=orange!40,
  thick,
  inner sep = 0,
  minimum size = \nodeSize]
\tikzstyle{unknownNode}=[
  circle,
  draw=blue!50,
  fill=blue!20,
  thick, dashed,
  inner sep = 0,
  minimum size = \nodeSize]

\tikzstyle{edgeStyle} = [
  very thick,
  dashed
]
\tikzstyle{pathStyle} = [
  ultra thick,
  rounded corners,
  cap = round,
  color = red!90,
  join = round,
]

\tikzset{dashdot/.style={dash pattern=on .4pt off 4pt on 4pt off 4pt}}

\tikzstyle{policyActionStyle} = [
  ultra thick,
  rounded corners,
  cap = round,
  color = gray!80,
  join = round,
  dashed,
]

\tikzstyle{policyChosenActionStyle} = [
  ultra thick,
  rounded corners,
  cap = round,
  color = blue!90,
  join = round,
  dashdot,
]

\tikzstyle{legendStyle} = [
  matrix,fill=blue!10,draw=blue,very thick
]

\tikzstyle{visibleWallStyle} = [
  very thick,
  rounded corners,
  join = round
]

\newcommand{\robotSize}{.42}
\newcommand{\drawRobot}[1] {
  \fill (#1) circle (\robotSize);
  \node[] (robotInternal) at (#1) {};
}

\newcommand{\labelRobot}{
  \node[font=\footnotesize, below = .05 of robotInternal] {Robot};
}

\newcommand{\drawStart}{
  \node[] at (0, -0.1) {Start};
}

\newcommand{\drawGrid}{
  \draw[help lines,xstep=1,ystep=1] (-9,-2) grid (0,5);
  \foreach \x in {-9,...,0} { \node [anchor=south] at (\x,5) {\x}; }
  \foreach \y in {-2,...,5} { \node [anchor=west] at (0,\y) {\y}; }

}

% Legend: Known/Unknown objects
\newcommand{\drawLandmarksLegend}{
  \node [legendStyle] (my matrix) at (-7,-2.5){
    \node[] {\textbf{Legend}}; \pgfmatrixnextcell  \\
    \node[left] {Known Object:}; \pgfmatrixnextcell
    \node[visibleLandmark, rectangle] {Landmark}; \\
    \node[left] {Unknown Object:}; \pgfmatrixnextcell
    \node[invisibleLandmark, rectangle] {Landmark}; \\
  };
}

% Legend: Visited/Frontier/unknown nodes
\newcommand{\drawGraphLegend}{
  \node [legendStyle] (my matrix) at (-7,-2.5){
    \node[] {\textbf{Legend}}; \pgfmatrixnextcell  \\
    \node[left] {Visited Vertex:}; \pgfmatrixnextcell
    \node[visibleNode] {}; \\
    \node[left] {Frontier Vertex:}; \pgfmatrixnextcell
    \node[frontierNode] {}; \\
  };
}

\newcommand{\tikzpicturescale}{.6}
\begin{figure}[t]
  \centering
  \begin{tikzpicture}[scale=\tikzpicturescale]
    \drawBeliefSpace
    \drawFeatureArrows
    \end{tikzpicture}
  \caption{Simplified illustration of computing feature moments in the space of hypothesized landmarks (in this case, two kitchens).
    To compute the features over a landmark distribution, we compute the features for each action across all hypothesized landmark samples, and aggregate them by computing moment statistics.}
  \label{figFeatureComputation}
\end{figure}
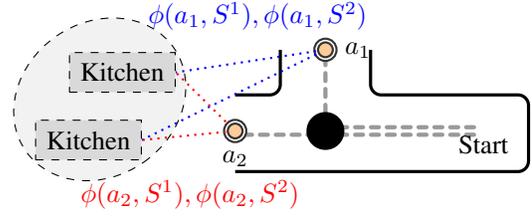

The vector $\phi (x, a, \Mapti)$ are features of the action and a \emph{single} landmark in $\Mapti$. It contains geometric features describing the shape of the action (e.g., the cumulative change in angle), the geometry of the landmark (e.g., the area of the landmark), and the relationship between the action and landmark (e.g., the difference between the ending and starting distances to the landmark). See~\cite{duvallet2013} for more details.

\subsection{Imitation Learning Formulation}

We use imitation learning to train the policy by treating action prediction as a multi-class classification problem:
given an expert demonstration, we wish to correctly predict their action among all possible actions for the same state.
Although prior work introduced imitation learning for training a direction following policy, it operated in partially known environments~\cite{duvallet2013}.
Instead, we train a belief space policy that reasons in a \emph{distribution} of hypothesized maps.

We assume the expert's policy $\pistar$ minimizes the unknown immediate
cost $C(x, a^*, \Mapt)$ of performing the demonstrated action $a^*$ from
state $x$, under the map distribution~$\Mapt$.
However, since we cannot directly observe the true costs of the expert's policy, we must instead minimize a surrogate loss that penalizes disagreements between the expert's action~$a^*$ and the policy's action~$a$, using the multi-class hinge loss~\cite{Crammer2002}:
\begin{equation}
\label{eqnSVMloss}
% Add some negative horizontal space to prevent overfull hbox
\ell \parens{x, \astar \!, c, \Mapt} \! =  \! \max \! \parens{ \! 0, 1 \! + \! \costMaptAstar \! - \! \min_{a \ne \astar} \brackets{\costMapt} \!}.
\end{equation}
The minimum of this loss occurs when the cost of the expert's action is lower than the cost of all other actions, with a margin of one. This loss can be re-written and combined with \autoref{eqnBeliefPolicyVectorK} to yield:
\begin{equation} \label{eqLossAugmentation}
  \ell \parens{x, \astar, W, \Mapt} = \Wtranspose F_\astar - \min_a \brackets{ \Wtranspose F_a - l_{xa} },
\end{equation}
where the margin $l_{xa} = 0$ if $a=\astar$ and $1$ otherwise.
This ensures that the expert's action is better than all other actions by a margin~\cite{Ratliff2006}.
Adding a regularization term $\lambda$ to~\autoref{eqLossAugmentation} yields our complete optimization loss:
\begin{equation} \label{eqnOptimizationLoss}
\ell \parens{x, \astar, W, \Mapt} \! = \! \frac{\lambda}{2} \Vert W \Vert^2 + \Wtranspose F_\astar - \min_{a} \left[ \Wtranspose F_a - l_{xa} \right].
\end{equation}

Although this loss function is convex, it is not differentiable.
However, we can optimize it efficiently by taking the subgradient of~\autoref{eqnOptimizationLoss} and computing action predictions for the loss-augmented policy~\cite{Ratliff2006}:
%
%\begin{subequations}
\begin{align}
 \dldwfrac & = \lambda W + F_\astar - F_{a'} \\
  a' & = \argmin{a} \left[ \Wtranspose F_a - l_{xa} \right].
\end{align}
Note that $a'$ (the best loss-augmented action) is simply the solution to our policy using a loss-augmented cost.  This leads to the update rule for the weights $W$:
\begin{equation} \label{eqnUpdateRule}
  W_{t+1} \gets W_t - \alpha \; \dldwfrac
\end{equation}
with a learning rate $\alpha \propto 1/t^\gamma$.
Intuitively, if the current policy disagrees with the expert's demonstration, \autoref{eqnUpdateRule} decreases the weight (and thus the cost) for the features of the demonstrated action $F_\astar$, and increases the weight for the features of the planned action $F_{a'}$.
If the policy produces actions that agree with the expert's demonstration, the update will only be for the regularization term.
As in our prior work, we train the policy using the \DAgger{} (Dataset Aggregation) algorithm~\cite{Ross2011}, which learns a policy by iterating between collecting data (using the current policy) and applying expert corrections on all states visited by the policy (using the expert's demonstrated policy).

% Recap
Treating direction following in the space of possible semantic maps as a problem of sequential decision making under uncertainty provides an efficient approximate solution to the belief space planning problem.
By using a kernel embedding of the distribution of features for a given action, our approach can learn a policy that reasons about the distribution of semantic maps.
%Using imitation learning for training the policy is simple, elegant, and requires no complex engineering of components or tuning of parameters.

% -*- mode:LaTeX; mode:visual-line; mode:flyspell; fill-column:75 -*-

% NOTE: This file is included in the planning file. Defines \label{figCrossValidation}
% \input{tikz/xvalidation_histogram}

\section{Results}
\label{sec:results}
We implemented the algorithm on our voice-commandable wheelchair
(Fig.~\ref{fig:go_to_kitchen_down_hallway}), which is equipped with three
forward-facing cameras with a collective field-of-view of 120 degrees, and
forward- and rearward-facing LIDARs. We set up an experiment in which the
wheelchair was placed in a lobby within MIT's Stata Center, with several
hallways, offices, and lab spaces, as well as a kitchen on the same
floor. As scene understanding is not the focus of this paper, we placed
AprilTag fiducials~\cite{Olson2011} to identify the existence and semantic
type of regions
in the environment. We trained the HDCG models from a parallel corpus of
54 fully-labeled examples.  We then directed the wheelchair to execute the
novel instruction ``go to the kitchen that is down the hallway.''

\begin{table}[t]
    \newcolumntype{S}{c@{\hskip 0.22in}}
    \centering
    \caption{Direction following efficiency on the robot}
    \begin{tabularx}{0.9\linewidth}{Scccc}
        \toprule
        & \multicolumn{2}{c}{Distance (m)} & \multicolumn{2}{c}{Time (s)}\\
        \cmidrule{2-3} \cmidrule{4-5}
        Algorithm & Mean & Std Dev & Mean & Std Dev\\
        \midrule
        Known Map & 13.10 & 0.67 & \hphantom{0}62.48 & 16.61\\
        With Language & 12.62 & 0.62 & 122.14 & 32.48\\
        Without Language & 24.91 & 13.55 & 210.35 & 97.73\\
        \bottomrule
    \end{tabularx} \label{tab:results_robot}
    \postTableSpace
\end{table}

\begin{table}[t]
    \newcolumntype{S}{c@{\hskip 0.22in}}
    \centering
    \caption{Direction following efficiency in simulation}
    \begin{tabularx}{0.9\linewidth}{Scccc}
        \toprule
        & \multicolumn{2}{c}{Distance (m)} & \multicolumn{2}{c}{Time (s)}\\
        \cmidrule{2-3} \cmidrule{4-5}
        Algorithm & Mean & Std Dev & Mean & Std Dev\\
        \midrule
        Known Map & 12.88 & 0.06 & 18.32 & \hphantom{0}3.54\\
        With Language & 16.64 & 6.84 & 82.78 & 10.56\\
        Without Language & 25.28 & 12.99 & 85.57 & 17.80\\
        \bottomrule
    \end{tabularx} \label{tab:results_sim}
    \postTableSpace
\end{table}

We compare our framework against two other methods. The first emulates the previous
state-of-the-art and uses a known map of the environment in order to
infer the actions consistent with the route
direction. The second assumes no prior knowledge of the environment
(as with ours) and opportunistically grounds the command in the map, but
does not use language to modify the map.
 We performed six experiments with our algorithm, three with the
known map method, and five with the method that does not use
language, all of which were successful (the robot reached the kitchen).
\autoref{tab:results_robot} compares the total distance traveled and
execution time for the three methods.
Our algorithm resulted in paths with lengths
close to those of the known map, and significantly outperformed the
method that did not use language. Our framework did
require significantly more time to follow the directions than the known
map case, due to the fact
that it repeats the three steps of the algorithm when new sensor data
arrives.  \autoref{fig:belief_world} shows a visualization of the semantic
maps over several time steps for one successful run on the robot.

We performed a similar evaluation in a simulated environment comprised of
an office, hallway, and kitchen. With the robot starting in the office, we
ran ten simulations of each method. As with the physical experiment,
our method resulted in an average length closer to that of the known map
case, but with a longer average run time (\autoref{tab:results_sim}).
%
%
%
%

% -*- mode:LaTex; mode:visual-line; mode:flyspell; fill-column:75-*-

% Both figures are in here.
% -*- mode:LaTex; mode:visual-line; mode:flyspell; fill-column:75-*-

\newcommand{\addBoxPlot}[3]{
  \addplot[
    mark=*,
    color=black!70,
    very thick,
    boxplot,
    every box/.style={
      fill=#2,  % fill color is second argument
    },
    every average/.style={%
      /tikz/mark=diamond*,
    },
  ]table[y = #3] {#1}
          [above]  % Draw low/median/high on right.
          node at (boxplot whisker cs:\boxplotvalue{lower whisker},1)
          {\small \pgfmathprintnumber{\boxplotvalue{lower whisker}}}
          node at (boxplot box cs: \boxplotvalue{median},1)
          {\small \pgfmathprintnumber{\boxplotvalue{median}}}
          node at (boxplot whisker cs:\boxplotvalue{upper whisker},1)
          {\small \pgfmathprintnumber{\boxplotvalue{upper whisker}}}
          ;
}

\newcommand{\drawAmbiguousDistanceBoxPlots}[1][tb]{
  \begin{figure}[#1]
    \centering
    \begin{tikzpicture}
      \begin{axis}[
          width=0.45\textwidth,
          height=0.2\textwidth,
          boxplot/variable width,
          boxplot/draw direction=x,
          axis x line*=bottom,
          axis y line=left,
          xlabel={Mean ending distance error (m)},
          xmin=0,
          xmax=15,
          xtick={0,5,10,15},
          ylabel={Belief space reasoning:},
          ytick={1,2},
          yticklabels={with, without},
          yticklabel style={rotate=90},
        ]
        \addBoxPlot{results/data/belief-distance.dat}{colorBelief}{validation};
        \addBoxPlot{results/data/nlmmp-distance.dat}{colorNLMMP}{validation};
      \end{axis}
    \end{tikzpicture}
  \caption
      {Tukey box plots showing the mean ending distance error of 27 natural language directions
        over 200 cross-validation trials, with and without belief space reasoning.
        Reasoning about the distribution of landmarks (with)
        improves direction following performance compared to without.}
    \label{figAmbiguousDistanceBoxPlots}
  \end{figure}
}

\definecolor{colorGreen}{RGB}{151, 205, 111}
\definecolor{colorOrange}{RGB}{247, 149, 89}
\definecolor{colorBlue}{RGB}{62, 175, 228}

\newcommand{\drawFullDirectionAmbiguousEight}[1][tb]{
  \begin{figure}[#1]
    \centering
    \begin{tikzpicture}
      \draw (0, 0) node[name=image, anchor=south west, inner sep=0pt, outer sep=0pt]{
        \includegraphics[
          % trim = left bottom right top
          trim = 60mm 50mm 84mm 72mm, clip,
          width = 0.4\textwidth
        ]{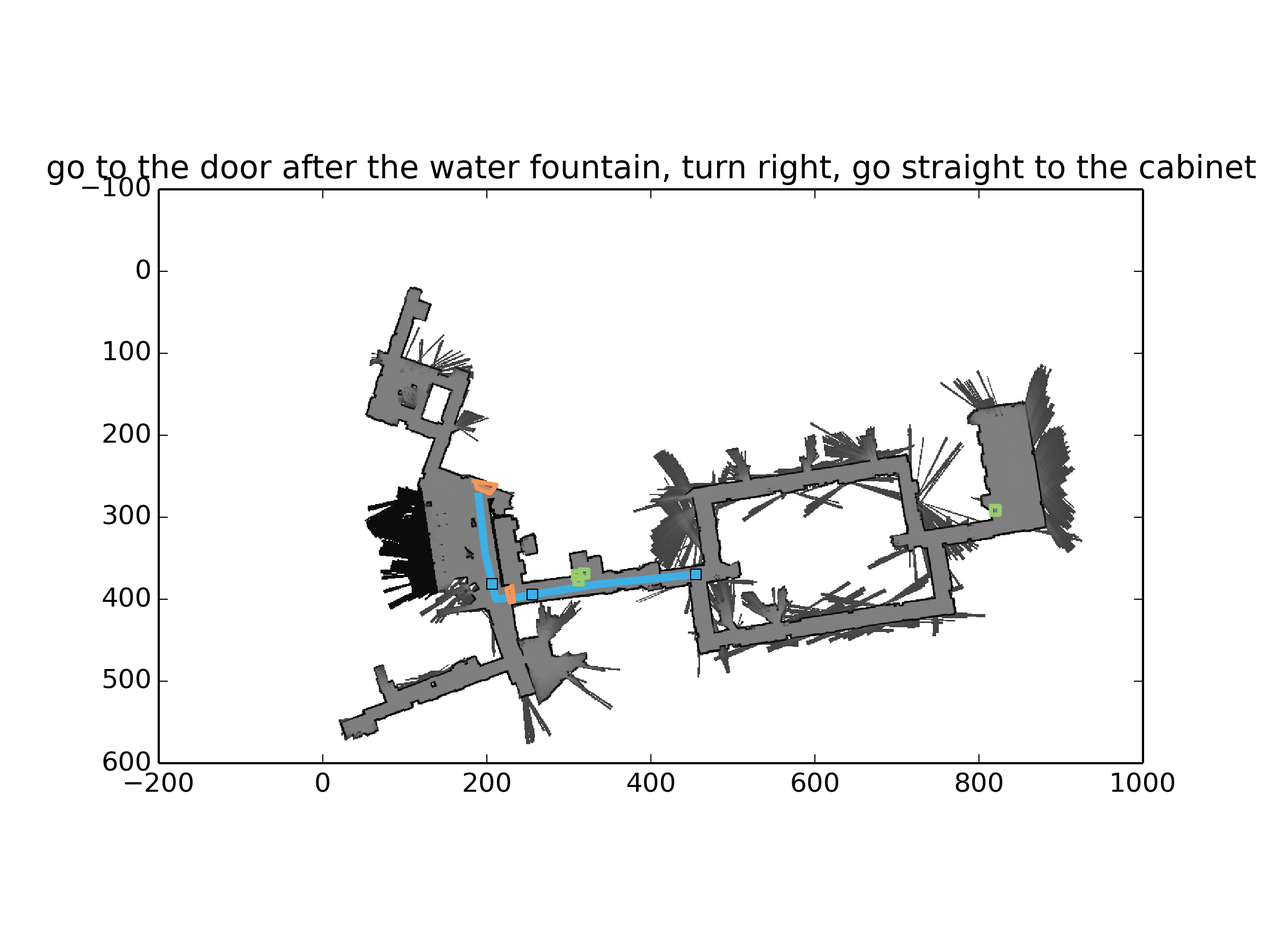}
      };
      \begin{scope}[x={(image.south east)},y={(image.north west)}]

        %% Grid
        %% \draw[help lines,xstep=.1,ystep=.1] (0,0) grid (1,1);
        %% \foreach \x in {0,1,...,9} { \node [anchor=north] at (\x/10,0) {0.\x}; }
        %% \foreach \y in {0,1,...,9} { \node [anchor=east] at (0,\y/10) {0.\y}; }

        \node[align=center,colorGreen, landmarkText] (fountains) at (0.55, 0.58) {Water\\ Fountains};
        \node[colorOrange, landmarkText] (door) at (0.35, 0.1) {Door};
        \node[colorOrange, landmarkText] (cabinet) at (0.45, 0.8) {Cabinet};

        \node[] (start) at (0.88, 0.35) {};
        \node[above left=18pt, landmarkText] (startText) at (start) {Start};
        \draw[->, colorBlue, ultra thick, shorten >= 5pt] (startText) -- (start);

      \end{scope}
    \end{tikzpicture}
    \caption{
      Ground truth path for the direction
      ``go to the door after the water fountain, turn right, go straight to
      the cabinet.'' The direction contains information about the door's location
      (i.e., it is after the water fountain) that is important to distinguishing
      it from the other doors in the same hallway.}
    \label{figFullDirectionAmbiguousEight}
  \end{figure}
}

\drawFullDirectionAmbiguousEight

\drawAmbiguousDistanceBoxPlots

% Goal:get bigger space of language
To evaluate the performance of the learned belief space policy in isolation on a larger corpus of natural language directions (with more verbs, spatial relations, and landmarks), we performed cross-validation trials of the policy operating in a simplified simulated map.
% Corpus description
We evaluated the policy using a corpus of~55 multi-step natural language directions, some of which refer to navigation landmarks (for example, the direction shown in \cref{figFullDirectionAmbiguousEight}).
These directions are similar to those in our prior work~\cite{duvallet2013}.
% Evaluate with cross-validation
For this cross-validation evaluation, we trained the policy on 28 randomly-sampled directions then evaluated the learned policy on the remaining 27 directions (measuring the average ending distance error across the  held out directions).
The results of this experiment, shown in \cref{figAmbiguousDistanceBoxPlots}, demonstrate the benefit of using the additional information available in the direction to infer a distribution of possible environment models.
By contrast, our prior approach (without belief space reasoning) ignores this information which results in larger ending distance errors.

% -*- mode:LaTeX; mode:visual-line; mode:flyspell; fill-column:75 -*-

\section{Conclusions}

Robots that can understand and follow natural language directions in
unknown environments are one step towards intuitive human-robot
interaction.
Reasoning about parts of the environment that have not yet
been detected would help enable seamless coordination in human-robot teams.

We have generalized our prior work to move beyond object-relative
navigation in small, open environments. The primary contributions of this
work include:
\begin{itemize}
\item a hierarchical framework that learns a compact
    probabilistic graphical model for language understanding;
\item a semantic map inference algorithm that hypothesizes the existence
    and location of spatially coherent regions in large environments; and
\item a belief space policy that reasons directly over
    the hypothesized map distribution and is trained based on expert
    demonstrations.
\end{itemize}
%
%
% to reason over large, spatially extended
% regions, and use
% imitation learning to train policies that can handle spatial constraints in
% the belief space.  Our approach decomposes the problem of following natural
% language directions as three coupled problems:
% \begin{itemize}
% \item Understanding the command to generate a set of annotations (facts about the world), as well as the desired action and path constraints.
% \item Generating a distribution of maps which is consistent with observations from the robot's sensors and the language annotations.
% \item Learning a belief space policy that can predict a sequence of actions under this distribution of maps.
% \end{itemize}
Together, these algorithms are integral to efficiently interpreting and
following natural language route directions in unknown, spatially extended,
and complex environments.
% Each of these is models is learned from a distribution of training data.
% The HDCG model infers annotations from
% the linguistic elements in the command that are treated as noisy sensor
% observations.  Using a Rao-Blackwellized particle filter, we efficiently
% maintain a distribution over possible semantic maps.  A novel belief space
% imitation learning technique enables us to train a policy that can reason
% about the map uncertainty and take a sequence of actions towards the goal.
% Each components updates as the robot travels.
%
We evaluated our algorithm through a series of simulations as well as
demonstrations on a voice-commandable autonomous wheelchair tasked with
following natural language route instructions in an office-like
environment.

% , using
% several simple commands. Our proposed formulation enables robots to
% efficiently follow directions through unknown environments, and
% allows us to separate the uncertainty in the language understanding,
% mapping, and policy. Furthermore, this decomposition may enable side-by-side
% comparison of different modules in the future.

In the future, we plan to carry out experiments on a more diverse set
of commands.  Other future work will focus on handling sequences of
commands, as well as streams of command that are given \emph{during}
execution to change the behavior of the robot.

% -*- mode:LaTeX; mode:visual-line; mode:flyspell; fill-column:75 -*-

\section*{Acknowledgments}

\footnotesize{
This work was supported in part by the Robotics Consortium of the U.S.~Army
Research Laboratory under the Collaborative Technology Alliance Program,
Cooperative Agreement W911NF-10-2-0016, and by ONR under MURI grant
``Reasoning in Reduced Information Spaces'' (no. N00014-09-1-1052).
}

\bibliographystyle{IEEEtran}
\bibliography{references}

\end{document}